\documentclass[11pt,letterpaper]{article}
\usepackage{emnlp2016}
\usepackage{times}
\usepackage{latexsym}

\usepackage{graphicx} 
\usepackage{subfig} 

\usepackage{algorithm}
\usepackage{algorithmic}

\usepackage{amsmath,amsfonts,amssymb}
\usepackage{mathtools}
\usepackage{multirow}
\usepackage{blkarray}
\usepackage{bm}

\usepackage{url}

\usepackage{xcolor}

\emnlpfinalcopy

\def\emnlppaperid{42}

\title{Generalizing and Hybridizing Count-based and Neural Language Models}

\author{Graham Neubig$^\dagger$ \and Chris Dyer$^\ddagger$ \\
        $^\dagger$Carnegie Mellon University, USA \\
        $^\ddagger$Google DeepMind, United Kingdom}

\date{}

\begin{document}

\maketitle

\begin{abstract}
Language models (LMs) are statistical models that calculate probabilities over sequences of words or other discrete symbols.
Currently two major paradigms for language modeling exist: count-based $n$-gram models, which have advantages of scalability and test-time speed, and neural LMs, which often achieve superior modeling performance.
We demonstrate how both varieties of models can be unified in a single modeling framework that defines a set of probability distributions over the vocabulary of words, and then dynamically calculates mixture weights over these distributions.
This formulation allows us to create novel hybrid models that combine the desirable features of count-based and neural LMs, and experiments demonstrate the advantages of these approaches.%
\footnote{
  Work was performed while GN was at the Nara Institute of Science and Technology and CD was at Carnegie Mellon University.
  Code and data to reproduce experiments is available at \url{http://github.com/neubig/modlm}
}
\end{abstract}

\section{Introduction}
\label{sec:intro}

Language models (LMs) are statistical models that, given a sentence $w_1^I := w_1,\ldots,w_I$, calculate its probability $P(w_1^I)$.
LMs are widely used in applications such as machine translation and speech recognition, and because of their broad applicability they have also been widely studied in the literature.
The most traditional and broadly used language modeling paradigm is that of count-based LMs, usually smoothed $n$-grams \cite{witten91zero,chen96smoothing}.
Recently, there has been a focus on LMs based on neural networks \cite{nakamura90wordcategory,bengio06nnlm,mikolov10rnnlm}, which have shown impressive improvements in performance over count-based LMs.
On the other hand, these neural LMs also come at the cost of increased computational complexity at both training and test time, and even the largest reported neural LMs \cite{chen15largevocab,williams15scaling} are trained on a fraction of the data of their count-based counterparts \cite{brants07largelanguagemodels}.

In this paper we focus on a class of LMs, which we will call \textit{mixture of distributions LMs} (MODLMs; \S\ref{sec:modlm}).
Specifically, we define MODLMs as all LMs that take the following form, calculating the probabilities of the next word in a sentence $w_i$ given preceding context $\bm{c}$ according to a mixture of several component probability distributions $P_k(w_i|\bm{c})$:
\begin{equation}
\label{eq:modlm}
P(w_i|\bm{c}) = \sum_{k=1}^{K} \lambda_k(\bm{c}) P_k(w_i|\bm{c}).
\end{equation}
Here, $\lambda_k(\bm{c})$ is a function that defines the mixture weights, with the constraint that $\sum_{k=1}^K \lambda_k(\bm{c}) = 1$ for all $\bm{c}$.
This form is not new in itself, and widely used both in the calculation of smoothing coefficients for $n$-gram LMs \cite{chen96smoothing}, and interpolation of LMs of various varieties \cite{jelinek80interpolation}.

The main contribution of this paper is to demonstrate that depending on our definition of $\bm{c}$, $\lambda_k(\bm{c})$, and $P_k(w_i|\bm{c})$, Eq.~\ref{eq:modlm} can be used to describe not only $n$-gram models, but also feed-forward \cite{nakamura90wordcategory,bengio06nnlm,schwenk07cslm} and recurrent \cite{mikolov10rnnlm,sundermeyer12lstmlm} neural network LMs (\S\ref{sec:existing}).
This observation is useful theoretically, as it provides a single mathematical framework that encompasses several widely used classes of LMs.
It is also useful practically, in that this new view of these traditional models allows us to create new models that combine the desirable features of $n$-gram and neural models, such as:
\begin{description}
\item[neurally interpolated $n$-gram LMs (\S\ref{sec:neuralngram}),] which learn the interpolation weights of $n$-gram models using neural networks, and
\item[neural/$n$-gram hybrid LMs (\S\ref{sec:hybridlm}),] which add a count-based $n$-gram component to neural models, allowing for flexibility to add large-scale external data sources to neural LMs.
\end{description}
We discuss learning methods for these models (\S\ref{sec:learningmodlm}) including a novel method of randomly dropping out more easy-to-learn distributions to prevent the parameters from falling into sub-optimal local minima.

Experiments on language modeling benchmarks (\S\ref{sec:experiments}) find that these models outperform baselines in terms of performance and convergence speed.

\section{Mixture of Distributions LMs}
\label{sec:modlm}

As mentioned above, MODLMs are LMs that take the form of Eq.~\ref{eq:modlm}.
This can be re-framed as the following matrix-vector multiplication:
\begin{equation*}
\bm{p}_{\bm{c}}^\intercal = D_{\bm{c}} \bm{\lambda}_{\bm{c}}^\intercal,
\end{equation*}
where $\bm{p}_{\bm{c}}$ is a vector with length equal to vocabulary size, in which the $j$th element $p_{\bm{c},j}$ corresponds to $P(w_i=j|\bm{c})$, $\bm{\lambda}_{\bm{c}}$ is a size $K$ vector that contains the mixture weights for the distributions, and $D_{\bm{c}}$ is a $J$-by-$K$ matrix, where element $d_{\bm{c},j,k}$ is equivalent to the probability $P_k(w_i = j | \bm{c})$.\footnote{We omit the subscript $\bm{c}$ when appropriate.}
An example of this formulation is shown in Fig.~\ref{fig:modlmmatrices}.

Note that all columns in $D$ represent probability distributions, and thus must sum to one over the $J$ words in the vocabulary, and that all $\bm{\lambda}$ must sum to 1 over the $K$ distributions.
Under this condition, the vector $\bm{p}$ will represent a well-formed probability distribution as well.
This conveniently allows us to calculate the probability of a single word $w_i=j$ by calculating the product of the $j$th row of $D_{\bm{c}}$ and $\bm{\lambda}_{\bm{c}}^\intercal$
\begin{equation*}
\label{eq:singleprob}
P_k(w_i = j | \bm{c}) = \bm{d}_{\bm{c},j}\bm{\lambda}_{\bm{c}}^\intercal.
\end{equation*}

In the sequel we show how this formulation can be used to describe several existing LMs (\S\ref{sec:existing}) as well as several novel model structures that are more powerful and general than these existing models (\S\ref{sec:novel}).

\begin{figure}
\centering
\small{
\[
\begin{blockarray}{ccccccccc}
\BAmulticolumn{4}{l}{\text{Probabilities }\bm{p}^\intercal} & & \BAmulticolumn{4}{r}{\text{Coefficients }\bm{\lambda}^\intercal} \\
 & \overbrace{\rule{0.5cm}{0pt}} & & & & & & \overbrace{\rule{0.5cm}{0pt}} & \\ \noalign{\vskip -1.5mm}
\begin{block}{c[c]c[cccc][c]c}
& p_{1}  & \multirow{4}{*}{=} & d_{1,1} & d_{1,2} & \cdots & d_{1,K} & \lambda_{1} & \\
& p_{2}  &                    & d_{2,1} & d_{2,2} & \cdots & d_{2,K} & \lambda_{2} & \\
& \vdots &                    & \vdots  & \vdots  & \ddots & \vdots  & \vdots      & \\
& p_{J}  &                    & d_{J,1} & d_{J,2} & \cdots & d_{J,K} & \lambda_{K} & \\
\end{block} \noalign{\vskip -1.5mm}
 & & & \BAmulticolumn{4}{c}{\underbrace{\rule{3.0cm}{0pt}}} & & \\
 & & & \BAmulticolumn{4}{c}{\text{Distribution matrix }D}   & & \\
\end{blockarray}
\]
}
\vspace{-7mm}\caption{\label{fig:modlmmatrices}MODLMs as linear equations}
\vspace{-2mm}
\end{figure}

\section{Existing LMs as Linear Mixtures}
\label{sec:existing}

\begin{figure*}[t!]
\begin{minipage}{.5\linewidth}
    \centering
    \subfloat[\label{fig:ngramasmodlm}Interpolated $n$-grams as MODLMs]{
\parbox{\textwidth}{
\centering
\small{
\[
\begin{blockarray}{ccccccccc}
\BAmulticolumn{4}{l}{\text{Probabilities }\bm{p}^\intercal} & \BAmulticolumn{5}{r}{\text{Heuristic interp. coefficients }\bm{\lambda}^\intercal} \\
 & \overbrace{\rule{0.5cm}{0pt}} & & & & & & \overbrace{\rule{0.5cm}{0pt}} & \\ \noalign{\vskip -1.5mm}
\begin{block}{c[c]c[cccc][c]c}
& p_{1}  & \multirow{4}{*}{=} & d_{1,1} & d_{1,2} & \cdots & d_{1,N} & \lambda_{1} & \\
& p_{2}  &                    & d_{2,1} & d_{2,2} & \cdots & d_{2,N} & \lambda_{2} & \\
& \vdots &                    & \vdots  & \vdots  & \ddots & \vdots  & \vdots      & \\
& p_{J}  &                    & d_{J,1} & d_{J,2} & \cdots & d_{J,N} & \lambda_{N} & \\
\end{block} \noalign{\vskip -1.5mm}
 & & & \BAmulticolumn{4}{c}{\underbrace{\rule{3.0cm}{0pt}}} & & \\
 \BAmulticolumn{9}{c}{\text{Count-based probabilities }P_{C}(w_{i}=j|w_{i-n+1}^{i-1})} \\
\end{blockarray}\vspace{-7mm}
\]
}
}
    }
\end{minipage}
\begin{minipage}{.5\linewidth}
    \subfloat[\label{fig:nnlmasmodlm}Neural LMs as MODLMs]{
\parbox{\textwidth}{
\centering
\small{
\[
\begin{blockarray}{ccccccccc}
\BAmulticolumn{4}{l}{\text{Probabilities }\bm{p}^\intercal} & \BAmulticolumn{5}{r}{\text{Result of softmax(NN(}\bm{c}\text{))}} \\
 & \overbrace{\rule{0.5cm}{0pt}} & & & & & & \overbrace{\rule{0.5cm}{0pt}} & \\ \noalign{\vskip -1.5mm}
\begin{block}{c[c]c[cccc][c]c}
& p_{1}  & \multirow{4}{*}{=} & 1      & 0      & \cdots & 0      & \lambda_{1} & \\
& p_{2}  &                    & 0      & 1      & \cdots & 0      & \lambda_{2} & \\
& \vdots &                    & \vdots & \vdots & \ddots & \vdots & \vdots      & \\
& p_{J}  &                    & 0      & 0      & \cdots & 1      & \lambda_{J} & \\
\end{block} \noalign{\vskip -1.5mm}
 & & & \BAmulticolumn{4}{c}{\underbrace{\rule{2.0cm}{0pt}}} & & \\
 & \BAmulticolumn{8}{c}{$J$\text{-by-}$J$\text{ identity matrix }I} \\
\end{blockarray}\vspace{-7mm}
\]
}
}
    }
\end{minipage}
    \caption{Interpretations of existing models as mixtures of distributions}
\vspace{-2mm}
\end{figure*}

\subsection{$n$-gram LMs as Mixtures of Distributions}
\label{sec:ngramasmodlm}

First, we discuss how count-based interpolated $n$-gram LMs fit within the MODLM framework.

\textbf{Maximum likelihood estimation:}
$n$-gram models predict the next word based on the previous $N$-1 words.
In other words, we set $\bm{c}=w_{i-N+1}^{i-1}$ and calculate $P(w_i|w_{i-N+1}^{i-1})$.
The maximum-likelihood (ML) estimate for this probability is
\begin{equation*}
P_{ML}(w_i|w_{i-N+1}^{i-1}) = c(w_{i-N+1}^i)/c(w_{i-N+1}^{i-1}),
\end{equation*}
where $c(\cdot)$ counts frequency in the training corpus.

\textbf{Interpolation:}
Because ML estimation assigns zero probability to word sequences where $c(w_{i-N+1}^i)=0$, $n$-gram models often interpolate the ML distributions for sequences of length 1 to $N$.
The simplest form is static interpolation
\begin{equation}
\label{eq:static}
P(w_i|w_{i-n+1}^{i-1}) = \sum_{n=1}^{N} \lambda_{S,n} P_{ML}(w_i|w_{i-n+1}^{i-1}).
\end{equation}
$\bm{\lambda}_{S}$ is a vector where $\lambda_{S,n}$ represents the weight put on the distribution $P_{ML}(w_i|w_{i-n+1}^{i-1})$.
This can be expressed as linear equations (Fig.~\ref{fig:ngramasmodlm}) by setting the $n$th column of $D$ to the ML distribution $P_{ML}(w_i|w_{i-n+1}^{i-1})$, and $\bm{\lambda}(\bm{c})$ equal to $\bm{\lambda}_{S}$.

Static interpolation can be improved by calculating $\bm{\lambda}(\bm{c})$ dynamically, using heuristics based on the frequency counts of the context \cite{good53population,katz87estimation,witten91zero}.
These methods define a context-sensitive fallback probability $\alpha(w_{i-n+1}^{i-1})$ for order $n$ models, and recursively calculate the probability of the higher order models from the lower order models:
\begin{multline}
\label{eq:fallback}
P(w_i|w_{i-n+1}^{i-1}) = \alpha(w_{i-n+1}^{i-1})P(w_i|w_{i-n+2}^{i-1})+ \\
                (1-\alpha(w_{i-n+1}^{i-1}))P_{ML}(w_i|w_{i-n+1}^{i-1}).
\end{multline}
%
To express this as a linear mixture, we convert $\alpha(w_{i-n+1}^{i-1})$ into the appropriate value for $\lambda_n(w_{i-N+1}^{i-1})$.
Specifically, the probability assigned to each $P_{ML}(w_i|w_{i-n+1}^{i-1})$ is set to the product of the fallbacks $\alpha$ for all higher orders and the probability of not falling back ($1-\alpha$) at the current level:
\begin{equation*}
\lambda_n(w_{i-N+1}^{i-1}) = (1-\alpha(w_{i-n+1}^{i-1}))\prod_{\tilde{n}=n+1}^{N} \alpha(w_{i-\tilde{n}+1}^{i-1}).
\end{equation*}

\textbf{Discounting:}
The widely used technique of discounting \cite{ney94structuring} defines a fixed discount $d$ and subtracts it from the count of each word before calculating probabilities:
\begin{equation*}
P_{D}(w_i|w_{i-n+1}^{i-1}) = (c(w_{i-n+1}^i) - d)/c(w_{i-n+1}^{i-1}).
\end{equation*}
Discounted LMs then assign the remaining probability mass after discounting as the fallback probability
\begin{align}
\beta_{D}(w_{i-n+1}^{i-1}) = & 1-\sum_{j=1}^J P_{D}(w_i=j|w_{i-n+1}^{i-1}), \nonumber \\
\label{eq:discountfallback}
P(w_i|w_{i-n+1}^{i-1}) = & \beta_{D}(w_{i-n+1}^{i-1})P(w_i|w_{i-n+2}^{i-1}) + \nonumber \\
                         &  P_{D}(w_i|w_{i-n+1}^{i-1}).
\end{align}
In this case, $P_{D}(\cdot)$ does not add to one, and thus violates the conditions for MODLMs stated in \S\ref{sec:modlm}, but it is easy to turn discounted LMs into interpolated LMs by normalizing the discounted distribution:
\begin{equation*}
P_{ND}(w_i|w_{i-n+1}^{i-1}) = \frac{P_{D}(w_i|w_{i-n+1}^{i-1})}{\sum_{j=1}^J P_{D}(w_i=j|w_{i-n+1}^{i-1})},
\end{equation*}
which allows us to replace $\beta(\cdot)$ for $\alpha(\cdot)$ and $P_{ND}(\cdot)$ for $P_{ML}(\cdot)$ in Eq.~\ref{eq:fallback}, and proceed as normal.

Kneser--Ney (KN; \newcite{kneser95improved}) and Modified KN \cite{chen96smoothing} smoothing further improve discounted LMs by adjusting the counts of lower-order distributions to more closely match their expectations as fallbacks for higher order distributions.
Modified KN is currently the de-facto standard in $n$-gram LMs despite occasional improvements \cite{teh06bayesiankn,durrett11lm}, and we will express it as $P_{KN}(\cdot)$.

\begin{figure*}[t!]
\begin{minipage}{.45\linewidth}
    \centering
    \subfloat[\label{fig:neuralngram}Neurally interpolated $n$-gram LMs]{
\parbox{\textwidth}{
\centering
\small{
\[
\begin{blockarray}{ccccccccc}
\BAmulticolumn{4}{l}{\text{Probabilities }\bm{p}^\intercal} & \BAmulticolumn{5}{r}{\underline{\text{Result of softmax(NN(}\bm{c}\text{))}}} \\
 & \overbrace{\rule{0.5cm}{0pt}} & & & & & & \overbrace{\rule{0.5cm}{0pt}} & \\ \noalign{\vskip -1.5mm}
\begin{block}{c[c]c[cccc][c]c}
& p_{1}  & \multirow{4}{*}{=} & d_{1,1} & d_{1,2} & \cdots & d_{1,N} & \lambda_{1} & \\
& p_{2}  &                    & d_{1,2} & d_{2,2} & \cdots & d_{2,N} & \lambda_{2} & \\
& \vdots &                    & \vdots  & \vdots  & \ddots & \vdots  & \vdots      & \\
& p_{J}  &                    & d_{J,1} & d_{J,2} & \cdots & d_{J,N} & \lambda_{N} & \\
\end{block} \noalign{\vskip -1.5mm}
 & & & \BAmulticolumn{4}{c}{\underbrace{\rule{3.0cm}{0pt}}} & & \\
 \BAmulticolumn{9}{c}{\text{Count-based probabilities }P_{C}(w_{i}=j|w_{i-n+1}^{i-1})} \\
\end{blockarray}\vspace{-7mm}
\]
}
}
    }
\end{minipage}
\begin{minipage}{.55\linewidth}
    \subfloat[\label{fig:hybridlm}Neural/$n$-gram hybrid LMs]{
\parbox{\textwidth}{
\centering
\small{
\[
\begin{blockarray}{ccccccccccc}
\BAmulticolumn{4}{l}{\text{Probabilities }\bm{p}^\intercal} & \BAmulticolumn{7}{r}{\text{Result of softmax(NN(}\bm{c}\text{))}} \\
 & \overbrace{\rule{0.5cm}{0pt}} & & & & & & & & \overbrace{\rule{0.5cm}{0pt}} & \\ \noalign{\vskip -1.5mm}
\begin{block}{c[c]c[cccccc][c]c}
& p_{1}  & \multirow{4}{*}{=} & d_{1,1} & \cdots & d_{1,N} & 1      & \cdots & 0      & \lambda_{1}   & \\
& p_{2}  &                    & d_{2,1} & \cdots & d_{2,N} & 0      & \cdots & 0      & \lambda_{2}   & \\
& \vdots &                    & \vdots  & \ddots & \vdots  & \vdots & \ddots & \vdots & \vdots        & \\
& p_{J}  &                    & d_{J,1} & \cdots & d_{J,N} & 0      & \cdots & 1      & \lambda_{J+N} & \\
\end{block} \noalign{\vskip -1.5mm}
 & & & \BAmulticolumn{6}{c}{\underbrace{\rule{4.0cm}{0pt}}} & & \\
 & \BAmulticolumn{10}{c}{\text{Count-based probabilities \underline{and} }J\text{-by-}J\text{ identity matrix}} \\
\end{blockarray}\vspace{-7mm}
\]
}
}
    }
\end{minipage}
    \caption{Two new expansions to $n$-gram and neural LMs made possible in the MODLM framework}
\vspace{-2mm}
\end{figure*}

\subsection{Neural LMs as Mixtures of Distributions}
\label{sec:nnlmasmodlm}

In this section we demonstrate how neural network LMs can also be viewed as an instantiation of the MODLM framework.

\textbf{Feed-forward neural network LMs:}
Feed-forward LMs \cite{bengio06nnlm,schwenk07cslm} are LMs that, like $n$-grams, calculate the probability of the next word based on the previous words.
Given context $w_{i-N+1}^{i-1}$, these words are converted into real-valued word representation vectors $\bm{r}_{i-N+1}^{i-1}$, which are concatenated into an overall representation vector $\bm{q} = \oplus(\bm{r}_{i-N+1}^{i-1})$, where $\oplus(\cdot)$ is the vector concatenation function.
$\bm{q}$ is then run through a series of affine transforms and non-linearities defined as function $\text{NN}(\bm{q})$ to obtain a vector $\bm{h}$.
For example, for a one-layer neural network with a tanh non-linearity we can define
\begin{equation}
\label{sec:tanhnn}
\text{NN}(\bm{q}) := \tanh(\bm{q}W_{q}+\bm{b}_{q}),
\end{equation}
where $W_{q}$ and $\bm{b}_{q}$ are weight matrix and bias vector parameters respectively.
Finally, the probability vector $\bm{p}$ is calculated using the softmax function $\bm{p} = \text{softmax}(\bm{h}W_{s}+\bm{b}_{s})$, similarly parameterized.

As these models are directly predicting $\bm{p}$ with no concept of mixture weights $\bm{\lambda}$, they cannot be interpreted as MODLMs as-is.
However, we can perform a trick shown in Fig.~\ref{fig:nnlmasmodlm}, not calculating $\bm{p}$ directly, but instead calculating mixture weights $\bm{\lambda} = \text{softmax}(\bm{h}W_{s}+\bm{b}_{s})$,
and defining the MODLM's distribution matrix $D$ as a $J$-by-$J$ identity matrix.
This is equivalent to defining a linear mixture of $J$ Kronecker $\delta_j$ distributions, the $j$th of which assigns a probability of 1 to word $j$ and zero to everything else, and estimating the mixture weights with a neural network.
While it may not be clear why it is useful to define neural LMs in this somewhat roundabout way, we describe in \S\ref{sec:novel} how this opens up possibilities for novel expansions to standard models.

\textbf{Recurrent neural network LMs:}
LMs using recurrent neural networks (RNNs) \cite{mikolov10rnnlm} consider not the previous few words, but also maintain a hidden state summarizing the sentence up until this point by re-defining the net in Eq.~\ref{sec:tanhnn} as
\begin{equation*}
\text{RNN}(\bm{q}_i) := \tanh(\bm{q}_iW_{q}+\bm{h}_{i-1}W_{h}+\bm{b}_{q}),
\end{equation*}
where $\bm{q}_i$ is the current input vector and $\bm{h}_{i-1}$ is the hidden vector at the previous time step.
This allows for consideration of long-distance dependencies beyond the scope of standard $n$-grams, and LMs using RNNs or long short-term memory (LSTM) networks \cite{sundermeyer12lstmlm} have posted large improvements over standard $n$-grams and feed-forward models.
Like feed-forward LMs, LMs using RNNs can be expressed as MODLMs by predicting $\bm{\lambda}$ instead of predicting $\bm{p}$ directly.

\section{Novel Applications of MODLMs}
\label{sec:novel}

This section describes how we can use this framework of MODLMs to design new varieties of LMs that combine the advantages of both $n$-gram and neural network LMs.

\subsection{Neurally Interpolated $n$-gram Models}
\label{sec:neuralngram}

The first novel instantiation of MODLMs that we propose is \textit{neurally interpolated $n$-gram models}, shown in Fig.~\ref{fig:neuralngram}.
In these models, we set $D$ to be the same matrix used in $n$-gram LMs, but calculate $\bm{\lambda}(\bm{c})$ using a neural network model.
As $\bm{\lambda}(\bm{c})$ is learned from data, this framework has the potential to allow us to learn more intelligent interpolation functions than the heuristics described in \S\ref{sec:ngramasmodlm}.
In addition, because the neural network only has to calculate a softmax over $N$ distributions instead of $J$ vocabulary words, training and test efficiency of these models can be expected to be much greater than that of standard neural network LMs.

Within this framework, there are several design decisions.
First, how we decide $D$: do we use the maximum likelihood estimate $P_{ML}$ or KN estimated distributions $P_{KN}$?
Second, what do we provide as input to the neural network to calculate the mixture weights?
To provide the neural net with the same information used by interpolation heuristics used in traditional LMs, we first calculate three features for each of the $N$ contexts $w_{i-n+1}^{i-1}$: a binary feature indicating whether the context has been observed in the training corpus ($c(w_{i-n+1}^{i-1})>0$), the log frequency of the context counts ($\log(c(w_{i-n+1}^{i-1}))$ or zero for unobserved contexts), and the log frequency of the number of unique words following the context ($\log(u(w_{i-n+1}^{i-1}))$ or likewise zero).
When using discounted distributions, we also use the log of the sum of the discounted counts as a feature.
We can also optionally use the word representation vector $\bm{q}$ used in neural LMs, allowing for richer representation of the input, but this may or may not be necessary in the face of the already informative count-based features.

\subsection{Neural/$n$-gram Hybrid Models}
\label{sec:hybridlm}

Our second novel model enabled by MODLMs is \textit{neural/$n$-gram hybrid models}, shown in Fig.~\ref{fig:hybridlm}.
These models are similar to neurally interpolated $n$-grams, but $D$ is augmented with $J$ additional columns representing the Kronecker $\delta_j$ distributions used in the standard neural LMs.
In this construction, $\boldsymbol{\lambda}$ is still a stochastic vector, but its contents are both the mixture coefficients for the count-based models and direct predictions of the probabilities of words.
Thus, the learned LM can use count-based models when they are deemed accurate, and deviate from them when deemed necessary.

This model is attractive conceptually for several reasons.
First, it has access to all information used by both neural and $n$-gram LMs, and should be able to perform as well or better than both models.
Second, the efficiently calculated $n$-gram counts are likely sufficient to capture many phenomena necessary for language modeling, allowing the neural component to focus on learning only the phenomena that are not well modeled by $n$-grams, requiring fewer parameters and less training time.
Third, it is possible to train $n$-grams from much larger amounts of data, and use these massive models to bootstrap learning of neural nets on smaller datasets.

\section{Learning Mixtures of Distributions}
\label{sec:learningmodlm}

While the MODLM formulations of standard heuristic $n$-gram LMs do not require learning, the remaining models are parameterized.
This section discusses the details of learning these parameters.

\subsection{Learning MODLMs}

The first step in learning parameters is defining our training objective.
Like most previous work on LMs \cite{bengio06nnlm}, we use a negative log-likelihood loss summed over words $w_i$ in every sentence $\bm{w}$ in corpus $\mathcal{W}$
\begin{equation*}
L(\mathcal{W}) = -\sum_{\bm{w} \in \mathcal{W}} \sum_{w_i \in \bm{w}} \log P(w_i|\bm{c}),
\end{equation*}
where $\bm{c}$ represents all words preceding $w_i$ in $\bm{w}$ that are used in the probability calculation.
As noted in Eq.~\ref{eq:singleprob}, $P(w_i=j|\bm{c})$ can be calculated efficiently from the distribution matrix $D_{\bm{c}}$ and mixture function output $\bm{\lambda}_{\bm{c}}$.

Given that we can calculate the log likelihood, the remaining parts of training are similar to training for standard neural network LMs.
As usual, we perform forward propagation to calculate the probabilities of all the words in the sentence, back-propagate the gradients through the computation graph, and perform some variant of stochastic gradient descent (SGD) to update the parameters.

\subsection{Block Dropout for Hybrid Models}
\label{sec:struct}

While the training method described in the previous section is similar to that of other neural network models, we make one important modification to the training process specifically tailored to the hybrid models of \S\ref{sec:hybridlm}.

This is motivated by our observation (detailed in \S\ref{sec:exp-hybridlm}) that the hybrid models, despite being strictly more expressive than the corresponding neural network LMs, were falling into poor local minima with higher training error than neural network LMs.
This is because at the very beginning of training, the count-based elements of the distribution matrix in Fig.~\ref{fig:hybridlm} are already good approximations of the target distribution, while the weights of the single-word $\delta_j$ distributions are not yet able to provide accurate probabilities.
Thus, the model learns to set the mixture proportions of the $\delta$ elements to near zero and rely mainly on the count-based $n$-gram distributions.

To encourage the model to use the $\delta$ mixture components, we adopt a method called \textit{block dropout} \cite{ammar16structuraldropout}.
In contrast to standard dropout \cite{srivastava14dropout}, which drops out single nodes or connections, block dropout randomly drops out entire subsets of network nodes.
In our case, we want to prevent the network from over-using the count-based $n$-gram distributions, so for a randomly selected portion of the training examples (here, 50\%) we disable all $n$-gram distributions and force the model to rely on only the $\delta$ distributions.
To do so, we zero out all elements in $\bm{\lambda}(\bm{c})$ that correspond to $n$-gram distributions, and re-normalize over the rest of the elements so they sum to one.

\subsection{Network and Training Details}

Finally, we note design details that were determined based on preliminary experiments. 

\textbf{Network structures:} We used both feed-forward networks with $\tanh$ non-linearities and LSTM \cite{hochreiter97lstm} networks. 
Most experiments used single-layer 200-node networks, and 400-node networks were used for experiments with larger training data. Word representations were the same size as the hidden layer. Larger and multi-layer networks did not yield improvements.

\textbf{Training:} 
We used ADAM \cite{kingma15adam} with a learning rate of 0.001, and minibatch sizes of 512 words.
This led to faster convergence than standard SGD, and more stable optimization than other update rules.
Models were evaluated every 500k-3M words, and the model with the best development likelihood was used.
In addition to the block dropout of \S\ref{sec:struct}, we used standard dropout with a rate of 0.5 for both feed-forward \cite{srivastava14dropout} and LSTM \cite{pham14dropout} nets in the neural LMs and neural/$n$-gram hybrids, but not in the neurally interpolated $n$-grams, where it resulted in slightly worse perplexities.

\textbf{Features:}
If parameters are learned on the data used to train count-based models, they will heavily over-fit and learn to trust the count-based distributions too much.
To prevent this, we performed 10-fold cross validation, calculating count-based elements of $D$ for each fold with counts trained on the other 9/10.
In addition, the count-based contextual features in \S\ref{sec:neuralngram} were normalized by subtracting the training set mean, which improved performance. 

\section{Experiments}
\label{sec:experiments}

\subsection{Experimental Setup}

In this section, we perform experiments to evaluate the neurally interpolated $n$-grams (\S\ref{sec:exp-neuralngram}) and neural/$n$-gram hybrids (\S\ref{sec:exp-hybridlm}), the ability of our models to take advantage of information from large data sets (\S\ref{sec:exp-bigdata}), and the relative performance compared to post-facto static interpolation of already-trained models (\S\ref{sec:exp-static}).
For the main experiments, we evaluate on two corpora: the Penn Treebank (PTB) data set prepared by \newcite{mikolov10rnnlm},\footnote{\url{http://rnnlm.org/simple-examples.tgz}} and the first 100k sentences in the English side of the ASPEC corpus \cite{nakazawa15wat}\footnote{\url{http://lotus.kuee.kyoto-u.ac.jp/ASPEC/}} (details in Tab.~\ref{tab:data}).
The PTB corpus uses the standard vocabulary of 10k words, and for the ASPEC corpus we use a vocabulary of the 20k most frequent words.
Our implementation is included as supplementary material.

\begin{table}
\centering
\caption{Data sizes for the PTB and ASPEC corpora.}\label{tab:data}
\small
\begin{tabular}{l|rr||l|rr}
PTB   & Sent  & Word & ASP   & Sent & Word \\ \hline\hline
train & 42k   & 890k & train & 100k & 2.1M \\ 
valid & 3.4k  & 70k  & valid & 1.8k & 45k  \\
test  & 3.8k  & 79k  & test  & 1.8k & 46k  \\
\end{tabular}
\end{table}


\subsection{Results for Neurally Interpolated $n$-grams}
\label{sec:exp-neuralngram}

First, we investigate the utility of neurally interpolated $n$-grams.
In all cases, we use a history of $N=5$ and test several different settings for the models:

\textbf{Estimation type:} $\bm{\lambda}(\bm{c})$ is calculated with heuristics (HEUR) or by the proposed method using feed-forward (FF), or LSTM nets.

\textbf{Distributions:} We compare $P_{ML}(\cdot)$ and $P_{KN}(\cdot)$. For heuristics, we use Witten-Bell for ML and the appropriate discounted probabilities for KN.

\textbf{Input features:} As input features for the neural network, we either use only the count-based features (C) or count-based features together with the word representation for the single previous word (CR).

\begin{table}
\centering
\caption{PTB/ASPEC perplexities for traditional heuristic (HEUR) and proposed neural net (FF or LSTM) interpolation methods using ML or KN distributions, and count (C) or count+word representation (CR) features.}\label{tab:exp-neuralngram}
\small
\begin{tabular}{l||c|c|c}
Dst./Ft. & HEUR          & FF            & LSTM                            \\ \hline\hline
ML/C     & 220.5/265.9   & 146.6/164.5 & 144.4/162.7                   \\ 
ML/CR    & -             & 145.7/163.9 & 142.6/158.4                   \\ \hline
KN/C     & 140.8/156.5 & 138.9/152.5 & 136.8/151.1                   \\
KN/CR    & -             & 136.9/153.0 & \textbf{135.2}/\textbf{149.1} \\
\end{tabular}
\end{table}

From the results shown in Tab.~\ref{tab:exp-neuralngram}, we can first see that when comparing models using the same set of input distributions, the neurally interpolated model outperforms corresponding heuristic methods.
We can also see that LSTMs have a slight advantage over FF nets, and models using word representations have a slight advantage over those that use only the count-based features.
Overall, the best model achieves a relative perplexity reduction of 4-5\% over KN models.
Interestingly, even when using simple ML distributions, the best neurally interpolated $n$-gram model nearly matches the heuristic KN method, demonstrating that the proposed model can automatically learn interpolation functions that are nearly as effective as carefully designed heuristics.%
\footnote{
Neurally interpolated $n$-grams are also more efficient than standard neural LMs, as mentioned in \S\ref{sec:neuralngram}.
While a standard LSTM LM calculated 1.4kw/s on the PTB data, the neurally interpolated models using LSTMs and FF nets calculated 11kw/s and 58kw/s respectively, only slightly inferior to 140kw/s of heuristic KN.
}

\subsection{Results for Neural/$n$-gram Hybrids}
\label{sec:exp-hybridlm}

In experiments with hybrid models, we test a neural/$n$-gram hybrid LM using LSTM networks with both Kronecker $\delta$ and KN smoothed 5-gram distributions, trained either with or without block dropout.
As our main baseline, we compare to LSTMs with only $\delta$ distributions, which have reported competitive numbers on the PTB data set \cite{zaremba14rnnregularization}.%
\footnote{
  Note that unlike this work, we opt to condition only on in-sentence context, not inter-sentential dependencies,
  as training through gradient calculations over sentences is more straightforward and because examining the effect of cross-boundary information is not central to the proposed method. 
  Thus our baseline numbers are not directly comparable (i.e. have higher perplexity) to previous reported results on this data, but we still feel that the comparison is appropriate.
}
We also report results for heuristically smoothed KN 5-gram models, and the best neurally interpolated $n$-grams from the previous section for reference.

\begin{table}
\centering
\caption{PTB/ASPEC perplexities for traditional KN (1) and LSTM LMs (2), neurally interpolated $n$-grams (3), and neural/$n$-gram hybrid models without (4) and with (5) block dropout.
}\label{tab:hybridresult}
\small
\begin{tabular}{lll||cc}
     & Dist.       & Interp.      & PPL               \\ \hline\hline
(1)  & KN          & HEUR         & 140.8/156.5 \\
(2)  & $\delta$    & LSTM         & 105.9/116.9 \\ \hline
(3)  & KN          & LSTM         & 135.2/149.1 \\ \hline
(4)  & KN,$\delta$ & LSTM -BlDO   & 108.4/130.4 \\
(5)  & KN,$\delta$ & LSTM +BlDO   & 95.3 /104.5 \\ 
\end{tabular}
\vspace{-3mm}
\end{table}

The results, shown in Tab.~\ref{tab:hybridresult}, demonstrate that similarly to previous research, LSTM LMs (2) achieve a large improvement in perplexity over $n$-gram models, and that the proposed neural/$n$-gram hybrid method (5) further reduces perplexity by 10-11\% relative over this strong baseline.

\begin{figure}[t]
\begin{center}
\centerline{\includegraphics[width=\columnwidth]{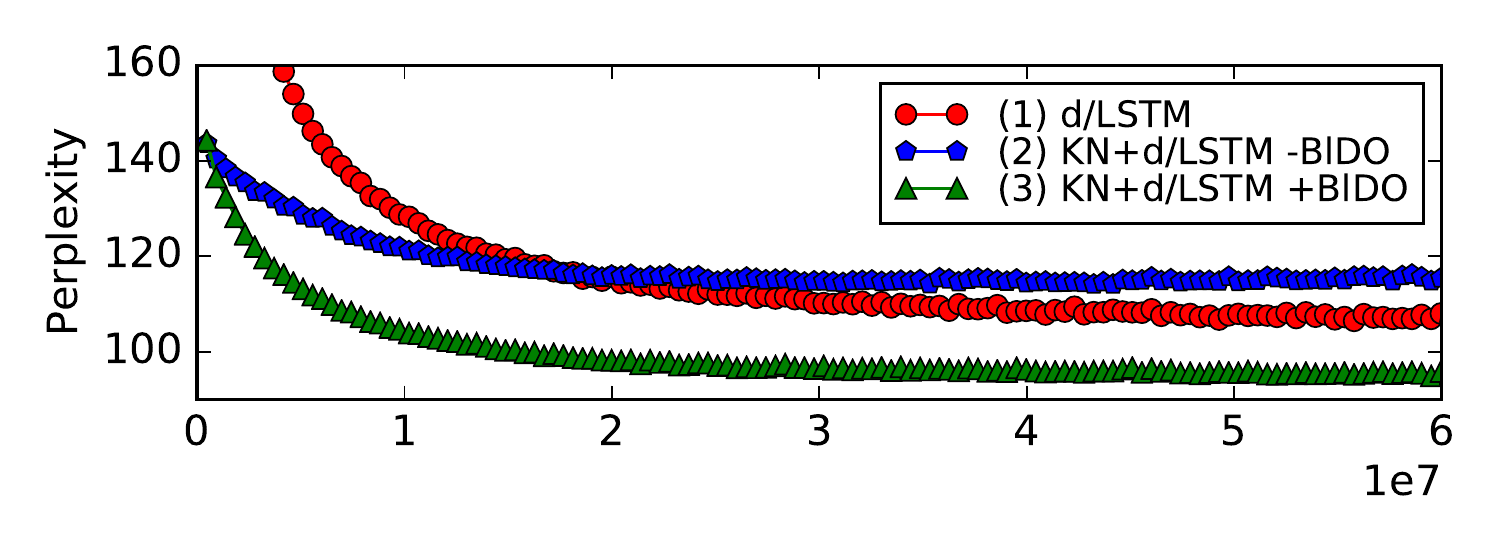}}
\vspace{-8.1mm}
\centerline{\includegraphics[width=\columnwidth]{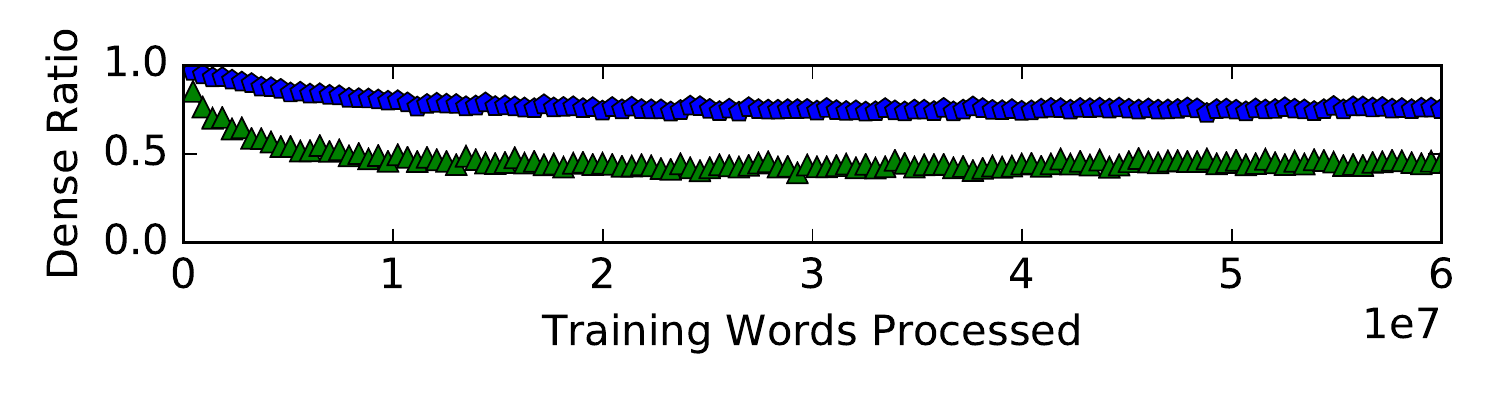}}
\vspace{-3mm}
\caption{Perplexity and dense distribution ratio of the baseline LSTM LM (1), and the hybrid method without (2) and with (3) block dropout.}
\label{fig:neural-convergence}
\end{center}
\vspace{-3mm}
\end{figure}

Comparing models without (4) and with (5) the proposed block dropout, we can see that this method contributes significantly to these gains.
To examine this more closely, we show the test perplexity for the three models using $\delta$ distributions in Tab.~\ref{fig:neural-convergence}, and the amount of the probability mass in $\bm{\lambda}(\bm{c})$ assigned to the non-$\delta$ distributions in the hybrid models.
From this, we can see that the model with block dropout quickly converges to a better result than the LSTM LM, but the model without converges to a worse result, assigning too much probability mass to the dense count-based distributions, demonstrating the learning problems mentioned in \S\ref{sec:struct}.

\begin{figure}[t]
\begin{center}
\centerline{\includegraphics[width=0.8\columnwidth]{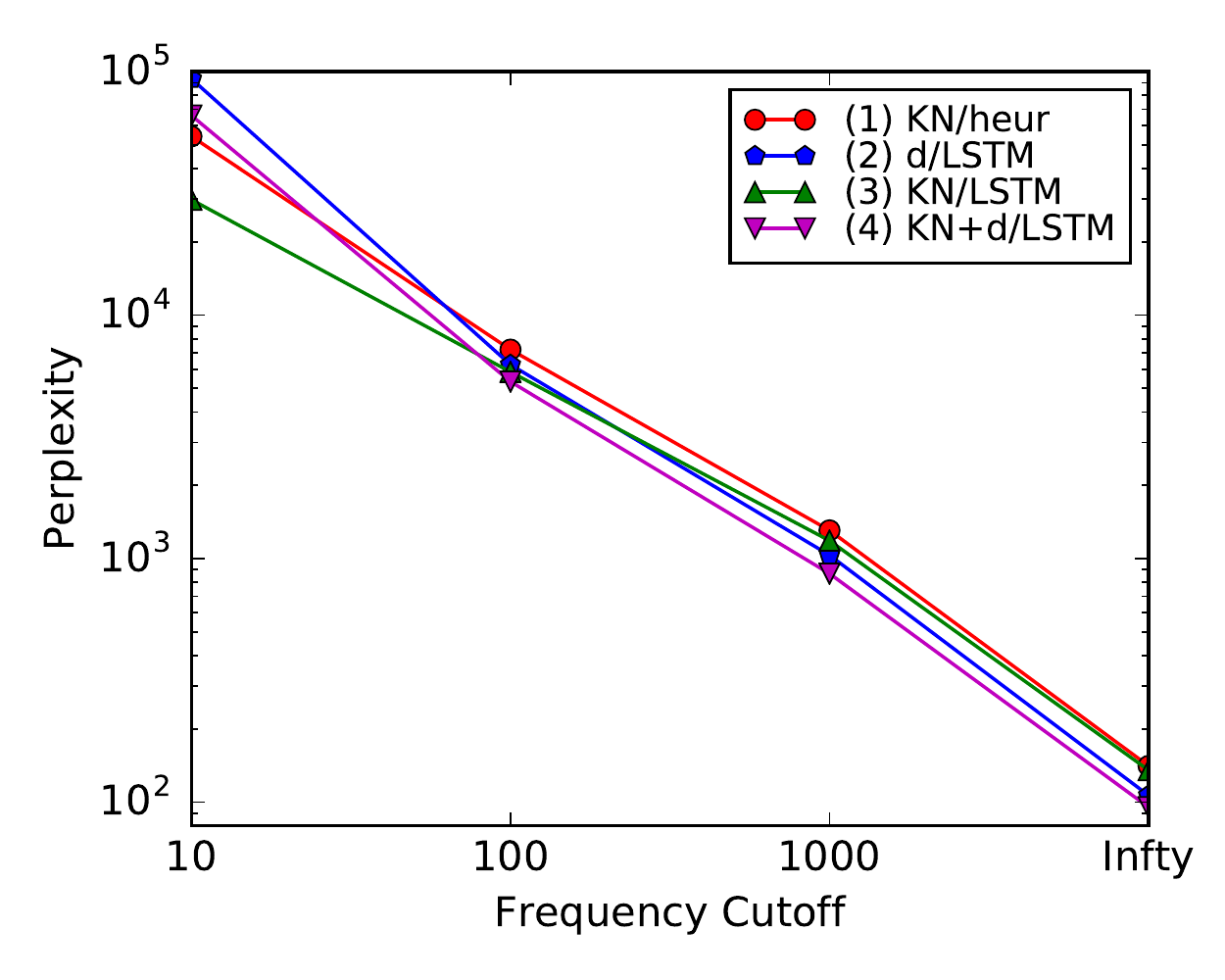}}
\vspace{-3mm}
\caption{Perplexities of (1) standard $n$-grams, (2) standard LSTMs, (3) neurally interpolated $n$-grams, and (4) neural/$n$-gram hybrids on lower frequency words.}
\label{fig:lowfreq}
\end{center}
\vspace{-3mm}
\end{figure} 

It is also of interest to examine exactly why the proposed model is doing better than the more standard methods.
One reason can be found in the behavior with regards to low-frequency words.
In Figure \ref{fig:lowfreq}, we show perplexities for words that appear $n$ times or less in the training corpus, for $n=10$, $n=100$, $n=1000$ and $n=\infty$ (all words).
From the results, we can first see that if we compare the baselines, LSTM language models achieve better perplexities overall but $n$-gram language models tend to perform better on low-frequency words, corroborating the observations of \newcite{chen15largevocab}.
The neurally interpolated $n$-gram models consistently outperform standard KN-smoothed $n$-grams, demonstrating their superiority within this model class.
In contrast, the neural/$n$-gram hybrid models tend to follow a pattern more similar to that of LSTM language models, similarly with consistently higher performance.

\subsection{Results for Larger Data Sets}
\label{sec:exp-bigdata}

To examine the ability of the hybrid models to use counts trained over larger amounts of data, we perform experiments using two larger data sets:

\textbf{WSJ:} The PTB uses data from the 1989 Wall Street Journal, so we add the remaining years between 1987 and 1994 (1.81M sents., 38.6M words).

\textbf{GW:} News data from the English Gigaword 5th Edition (LDC2011T07, 59M sents., 1.76G words).

We incorporate this data either by training net parameters over the whole large data, or by separately training count-based $n$-grams on each of PTB, WSJ, and GW, and learning net parameters on only PTB data.
The former has the advantage of training the net on much larger data.
The latter has two main advantages: 1) when the smaller data is of a particular domain the mixture weights can be learned to match this in-domain data; 2) distributions can be trained on data such as Google $n$-grams (LDC2006T13), which contain $n$-gram counts but not full sentences.

\begin{figure}[t]
\begin{center}
\centerline{\includegraphics[width=\columnwidth]{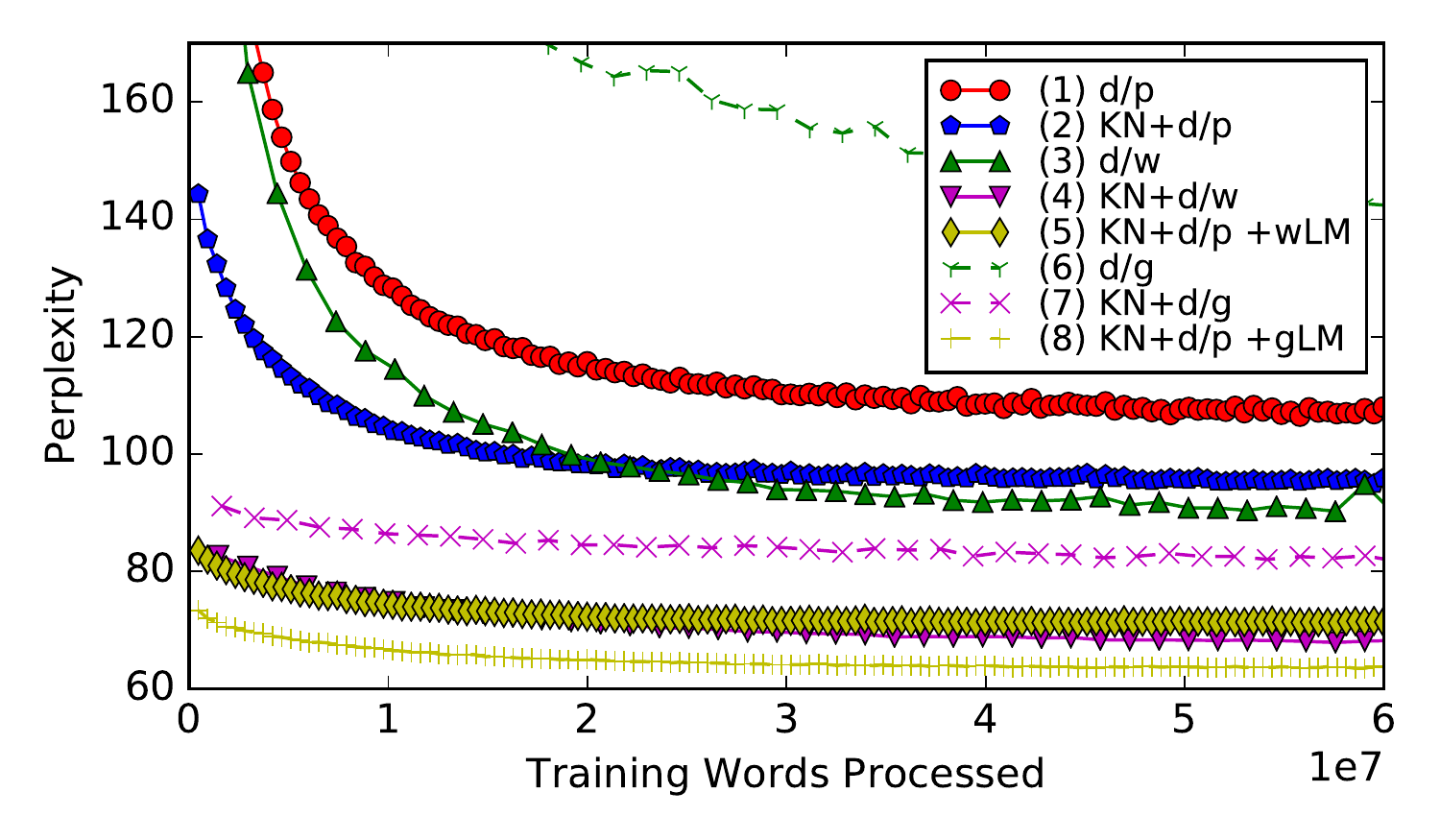}}
\vspace{-3mm}
\caption{Models trained on PTB (1,2), PTB+WSJ (3,4,5) or PTB+WSJ+GW (6,7,8) using standard neural LMs (1,3,6), neural/$n$-gram hybrids trained all data (2,4,7), or hybrids trained on PTB with additional $n$-gram distributions (5,8).}
\label{fig:big-data}
\end{center}
\vspace{-3mm}
\end{figure} 

In the results of Fig.~\ref{fig:big-data}, we can first see that the neural/$n$-gram hybrids significantly outperform the traditional neural LMs in the scenario with larger data as well.
Comparing the two methods for incorporating larger data, we can see that the results are mixed depending on the type and size of the data being used.
For the WSJ data, training on all data slightly outperforms the method of adding distributions, but when the GW data is added this trend reverses.
This can be explained by the fact that the GW data differs from the PTB test data, and thus the effect of choosing domain-specific interpolation coefficients was more prominent.

\subsection{Comparison with Static Interpolation}
\label{sec:exp-static}

Finally, because the proposed neural/$n$-gram hybrid models combine the advantages of neural and $n$-gram models, we compare with the more standard method of training models independently and combining them with static interpolation weights tuned on the validation set using the EM algorithm.
Tab.~\ref{tab:exp-interp} shows perplexities for combinations of a standard neural model (or $\delta$ distributions) trained on PTB, and count based distributions trained on PTB, WSJ, and GW are added one-by-one using the standard static and proposed LSTM interpolation methods.
From the results, we can see that when only PTB data is used, the methods have similar results, but with the more diverse data sets the proposed method edges out its static counterpart.%
\footnote{In addition to better perplexities, neural/$n$-gram hybrids are trained in a single pass instead of performing post-facto interpolation, which may give advantages when training for other objectives \cite{auli14decoderintegration,li15diversity}.}

\begin{table}
\centering
\caption{PTB perplexity for interpolation between neural ($\delta$) LMs and count-based models.}\label{tab:exp-interp}
\small
\begin{tabular}{l||c|c|c}
Interp & $\delta$+PTB  & +WSJ          & +GW           \\ \hline\hline
Lin.   & \textbf{95.1} & 70.5          & 65.8          \\ 
LSTM   & 95.3          & \textbf{68.3} & \textbf{63.5} \\
\end{tabular}
\end{table}


\section{Related Work}

A number of alternative methods focus on interpolating LMs of multiple varieties such as in-domain and out-of-domain LMs \cite{bulyko03classdependentmixtures,bacchiani06map,gulcehre15monolingual}.
Perhaps most relevant is \newcite{hsu07generalized}'s work on learning to interpolate multiple LMs using log-linear models.
This differs from our work in that it learns functions to estimate the fallback probabilities $\alpha_n(\bm{c})$ in Eq.~\ref{eq:fallback} instead of $\bm{\lambda}(\bm{c})$, and does not cover interpolation of $n$-gram components, non-linearities, or the connection with neural network LMs.
Also conceptually similar is work on adaptation of $n$-gram LMs, which start with $n$-gram probabilities \cite{dellapietra92adaptive,kneser93dynamicadaptation,rosenfeld96maximument,iyer99topicmixtures} and adapt them based on the distribution of the current document, albeit in a linear model.
There has also been work incorporating binary $n$-gram features into neural language models, which allows for more direct learning of $n$-gram weights \cite{mikolov11strategies}, but does not afford many of the advantages of the proposed model such as the incorporation of count-based probability estimates.
Finally, recent works have compared $n$-gram and neural models, finding that neural models often perform better in perplexity, but $n$-grams have their own advantages such as effectiveness in extrinsic tasks \cite{baltescu15pragmatic} and better modeling of rare words \cite{chen15largevocab}.

\section{Conclusion and Future Work}

In this paper, we proposed a framework for language modeling that generalizes both neural network and count-based $n$-gram LMs.
This allowed us to learn more effective interpolation functions for count-based $n$-grams, and to create neural LMs that incorporate information from count-based models.

As the framework discussed here is general, it is also possible that they could be used in other tasks that perform sequential prediction of words such as neural machine translation \cite{sutskever14sequencetosequence} or dialog response generation \cite{sordoni15nnresponse}.
In addition, given the positive results using block dropout for hybrid models, we plan to develop more effective learning methods for mixtures of sparse and dense distributions.

\section*{Acknowledgements}

We thank Kevin Duh, Austin Matthews, Shinji Watanabe, and anonymous reviewers for valuable comments on earlier drafts.
This work was supported in part by JSPS KAKENHI Grant Number 16H05873, and the Program for Advancing Strategic International Networks to Accelerate the Circulation of Talented Researchers.

\bibliography{myabbrv,main}
\bibliographystyle{emnlp2016}

\end{document}